\documentclass[conference,english]{IEEEtran}
\usepackage[T1]{fontenc}
\usepackage[latin9]{inputenc}
\usepackage{array}
\usepackage{url}
\usepackage{amsmath}
\usepackage{color}
\usepackage{graphicx}

\makeatletter

\usepackage{times}

\usepackage{epsfig}

\usepackage{babel}
\makeatother

\begin{document}

\title{Spatio-activity based object detection}

\author{Jarrad Springett and Jeroen Vendrig\\
Canon Information Systems Research Australia\\
{\tt\small \{jarrad,jeroen\}@research.canon.com.au}}

\maketitle


\begin{abstract}
We present the SAMMI lightweight object detection method which has
a high level of accuracy and robustness, and which is able to operate
in an environment with a large number of cameras. Background modeling
is based on DCT coefficients provided by cameras. Foreground detection
uses similarity in temporal characteristics of adjacent blocks of
pixels, which is a computationally inexpensive way to make use of
object coherence. Scene model updating uses the approximated median
method for improved performance. Evaluation at pixel level and application
level shows that SAMMI object detection performs better and faster
than the conventional Mixture of Gaussians method.
\end{abstract}

\section{Introduction}

The field of object detection has matured to the extent that its results
are sufficiently robust to be deployed in the real world. Systems
using object detection assist in mitigating security threats and gathering
marketing information. Research focus has shifted to further analysis
making use of object detection results, such as behavior analysis.
However, with the large number of surveillance cameras installed each
year, scalability of analysis solutions becomes a problem when one
server is connected to many cameras. Performing real-time video analysis
is not possible without investing significantly in extra infrastructure,
such as additional servers or dedicated analysis boxes.

As much of the resource usage in a video analysis is spent on the
underlying object detection, it makes sense to revisit detection techniques
and make them more suitable for deployment in large operations. Object
detection, and especially foreground/background separation, is the
bottle neck in an analysis system because the foreground separation
needed for object detection is performed on every pixel in the frame.
After objects have been found, video analysis can focus at the higher
level of objects. Obviously, there are many less objects in the average
frame than there are pixels. The problem could be addressed by reducing
the sampling. For example, analysis could be done at a lower frame
rate than the camera's capture rate, or processing could be performed
on a selected region of interest only. However, such approaches are
just shifting the problem and do not make optimal use of the information
provided by the cameras. 

In this paper, we present a lightweight object detection technique
that still has a high level of accuracy and robustness. In the context
of this paper, we assume that the camera has a fixed field of view.
Yet we do expect that an object detection method reinitialize quickly
when the field of view changes in the case of Pan-Tilt-Zoom (PTZ)
cameras. Also, we consider objects in the scene that are non-transient
during an entire recording, e.g. a parked car, to be part of the background.
Transient objects are considered foreground. A foreground object may
be stationary for part of the recording, while the background may
contain movement, e.g. a swaying tree. 

The paper is organized as follows. In section \ref{sec:Previous-work},
previous work in the field is described. In section \ref{sec:SAMMI-Ssystem-overview},
a general overview of the system and context in which the spatio-activity
based object detection operates is given. In section \ref{sec:System-details},
we present the details of our SAMMI (Spatio-Activity Multi-Mode with
Iterations) method. Finally, in sections \ref{sec:Evaluation} and
\ref{sec:Conclusions}, we evaluate the method and draw conclusions.

\section{\label{sec:Previous-work}Previous work}

The general concept of background modeling that is the basis for our
approach is well known in literature. The basic assumption is that
background is static and doesn't change, and that anything new must
be foreground. Early approaches, e.g. \cite{Nakanishi92}, update
a reference frame to be robust to changes in the scene or the appearance
of the scene, e.g. lighting changes, and do background subtraction
with incoming frames. Background modeling was popularized by Stauffer
and Grimson's Mixture of Gaussians approach \cite{Stauffer00}. The
Mixture of Gaussians approach does not maintain one model per pixel,
but several models in the form of a Gaussian. The models are not statically
classified as foreground or background, but are dynamically classified
after updating the model with the information from a new frame. More
than one model can be background, allowing for modeling of multi-modal
backgrounds, such as swaying trees.

Li and Sezan \cite{Li01} have used information about the neighbors
of pixels being classified in order to address the problem of spatial
and temporal aliasing on cameras. They assume that the probability
that a pixel is background depends on the classification of neighboring
pixels. The disadvantage of this approach is that it results in fragmentation
and erosion of foreground objects. For this reason, Li and Sezan apply
the neighbor based probability classification only after finding large
connected components. This approach prevents undesired hole filling,
but as a trade-off it rejects small foreground objects. The trade-off
may be acceptable in the background replacement application with close-up
and medium shots that Li and Sezan are targeting. But the loss of
small objects is not desirable in security applications, especially
if objects are small because of their distance to the camera. The
ability to detect objects with a greater range in sizes requires less
zoom to successfully detect objects, thus fewer cameras need be installed
to cover the same field of view \cite{Stauffer05}. The approach of
Li and Sezan shows that spatial proximity of pixels assists in classifying
them as foreground or background, but it is not sufficient without
defining a further relationship between the pixels. 

The most obvious relationship between pixels is based on the visual
characteristics of the pixels, such as color. Such relationships are
complex, e.g. because of texture, and also computationally expensive.
This approach depends very much on the progress in still image segmentation.

\section{\label{sec:SAMMI-Ssystem-overview}SAMMI system overview}

\begin{figure}
\includegraphics[bb=175bp 260bp 425bp 585bp,clip,width=0.8\columnwidth]{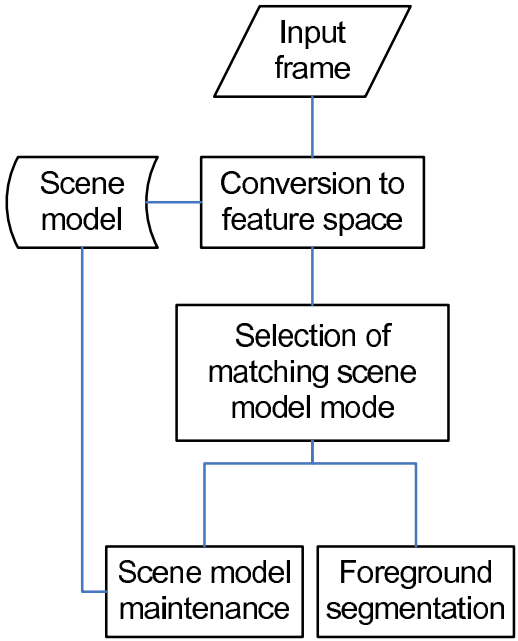}

\caption{Block diagram of the SAMMI system.\label{fig:overview}}

\end{figure}

The SAMMI (Spatio-Activity Multi-Mode with Iterations) method presented
in this paper, makes use of temporal characteristics to define the
relationship between neighboring pixels, or blocks of pixels in our
particular implementation. The temporal characteristics are cheap
to maintain and compare from a resource usage perspective, yet they
are effective because they make use of the temporal coherence of moving
objects. In this section,
we present a general overview of the system, as depicted in figure
\ref{fig:overview}. The key components are discussed in more detail
in section \ref{sec:System-details}. 

The system context has significant impact on the SAMMI algorithm.
The starting point is not raw image data, but a Motion JPEG stream
(intra-frame coding only) captured by a camera and transmitted over
a network. A departure from conventional approaches is that we don't
decompress the JPEG information to pixel-level information in the
spatial domain. Instead, processing is done on Discrete Cosine Transformation
(DCT) coefficients in the frequency domain and, as a consequence,
at 8x8 pixels block-level. The primary motivation for using the DCT
coefficients is the computational cost of the inverse transform to
the pixel domain. In addition, it allows for reduction in memory usage
and processing for scene modeling. Although storing 192 DCT coefficient
values is no different from storing 192 RGB pixels for an 8x8 block,
the information carried by the DCT coefficients is concentrated in
just a few coefficients. We found that using 8 of the 192 coefficients
is sufficient for acceptable detection accuracy. The disadvantage
of the block-level approach is that object edges are less accurate
than in a pixel-level approach. The impact of the blockiness of edges
depends on the application, e.g. security applications are invariant
under blockiness, and on the camera resolution.

The original input to our system is in the YCbCr color space. Principal
component analysis of the YCbCr component values of typical images
shows that Cb and Cr are more highly correlated to each other than
the alternative in-phase (I) and quadrature (Q) chrominance components
in the YIQ color space \cite{Pratt91}. Improving the independence
of DCT components improves the results of the DCT classifier described
in section \ref{sub:DCT-classifier}, allowing for greater invariance
under changing lighting conditions such as shadows. The first 6 Y
DCT coefficients are used in conjunction with the I and Q DC coefficients.

Using the input coefficients, the scene is modeled on a per-block
basis. For each block, several mode models are maintained, each representing
a state of the image, representing stationary objects and the underlying
background simultaneously. Mode models contain DCT coefficient values
and temporal information. This temporal information allows the age
and frequency of encountering a mode to be determined. For each mode,
the following temporal characteristics are recorded:

\begin{itemize}
\item Creation frame: the frame in which the mode first was created. We
refer to the difference between the current frame and the creation
frame as the mode age.
\item Hit count: the number of times that the mode has been matched. This
characteristic represents the activity of the mode.
\item Last matched frame: the frame that the mode was last matched in. 
\item Potential removal frame: the frame in which a mode is scheduled for
removal in, if no further matches to this mode occur.
\end{itemize}
The best mode match for each block in the image is determined by a
DCT classifier. The DCT classifier also determines if a new mode is
created for a given DCT block. A spatial classifier, described in
more detail in section \ref{sub:Spatial-classifier}, is then applied
to improve the selection of matching modes based on the temporal information
in adjacent modes.

The temporal information and DCT coefficient values of matched modes
are updated using an approximated median filter as described in section
\ref{sub:Approximated-median-filter}.

\begin{figure}
\begin{tabular}{cc}
\includegraphics[width=0.45\columnwidth]{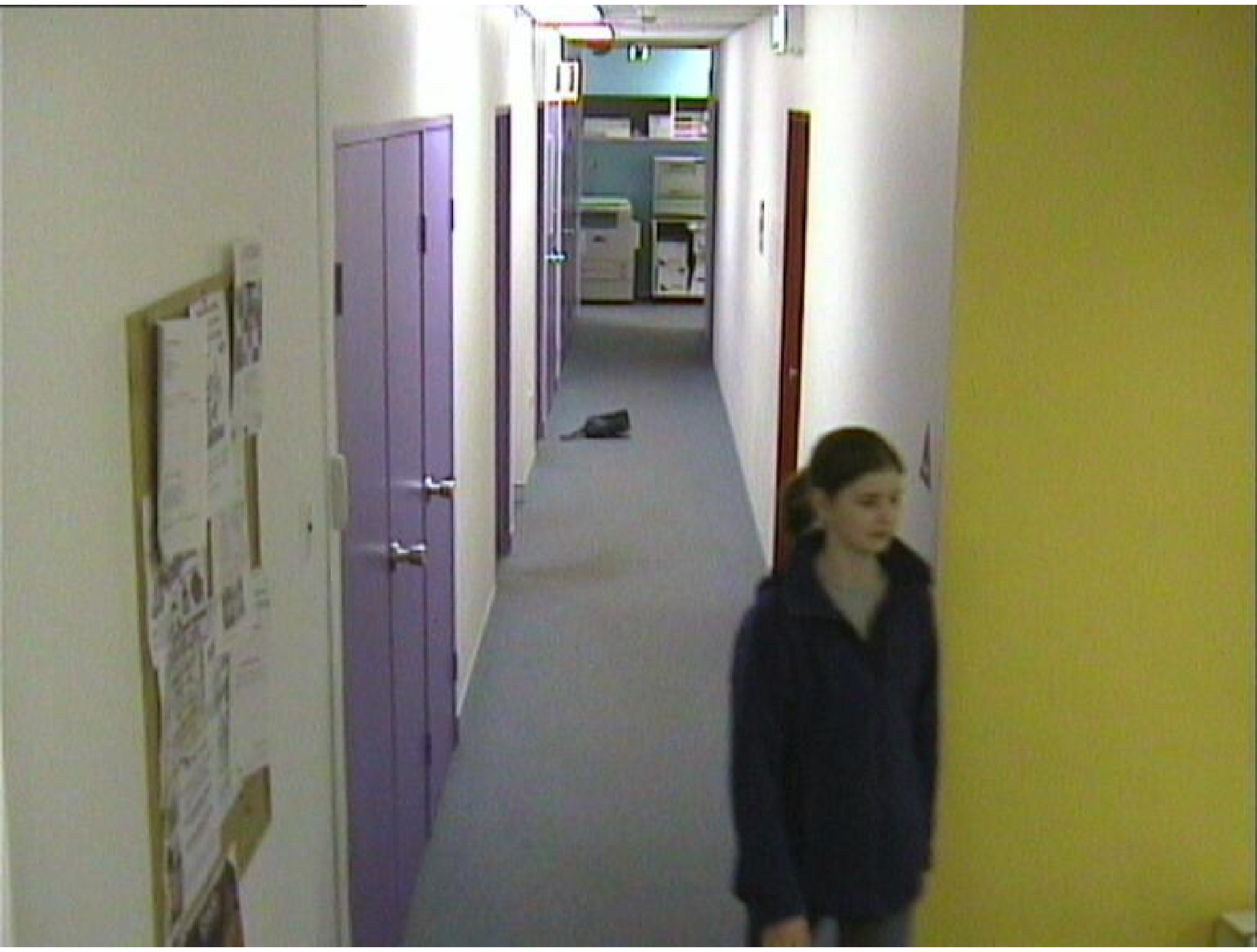} & \includegraphics[width=0.45\columnwidth]{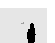}\tabularnewline
\end{tabular}

\caption{Detecting an abandoned bag: Input image (left), Ages of matched modes
in a sequence (right)\label{fig:Age image}}

\end{figure}

The output of this system is a set of connected components that represent
foreground objects. Flood fill connected component analysis is used,
using the creation time of the matched mode for each block as the
merging criterion. A block is connected to an adjacent block if and
only if the creation frame of both matched mode models is either greater
than a threshold value T, or smaller or equal to the threshold value
T. Therefore, each connected component will be exclusively comprised
of blocks that are either entirely of age greater than T or less than
or equal to T. Figure \ref{fig:Age image} shows an example where
a dropped bag has an intermediate age, compared to the person and
the background. 

\section{\label{sec:System-details}SAMMI method}

\subsection{\label{sub:DCT-classifier}DCT classifier}

Each block in each input image is compared to the mode models for
that block in order to give the probability that the input image block
matches the mode model. A high probability of a match means that the
input is likely to have been encountered before. If the highest probability
mode match for a block is below a threshold T, the input block is
likely to have not been encountered previously and a new mode is created.
The highest probability mode match can be used directly for creating
blob output. However,
we employ further processing steps as described in sections \ref{sub:Spatial-classifier}
and \ref{sub:Active-mode-bonus}.

\begin{figure*}
\begin{tabular}{ccc}
\includegraphics[width=0.3\textwidth]{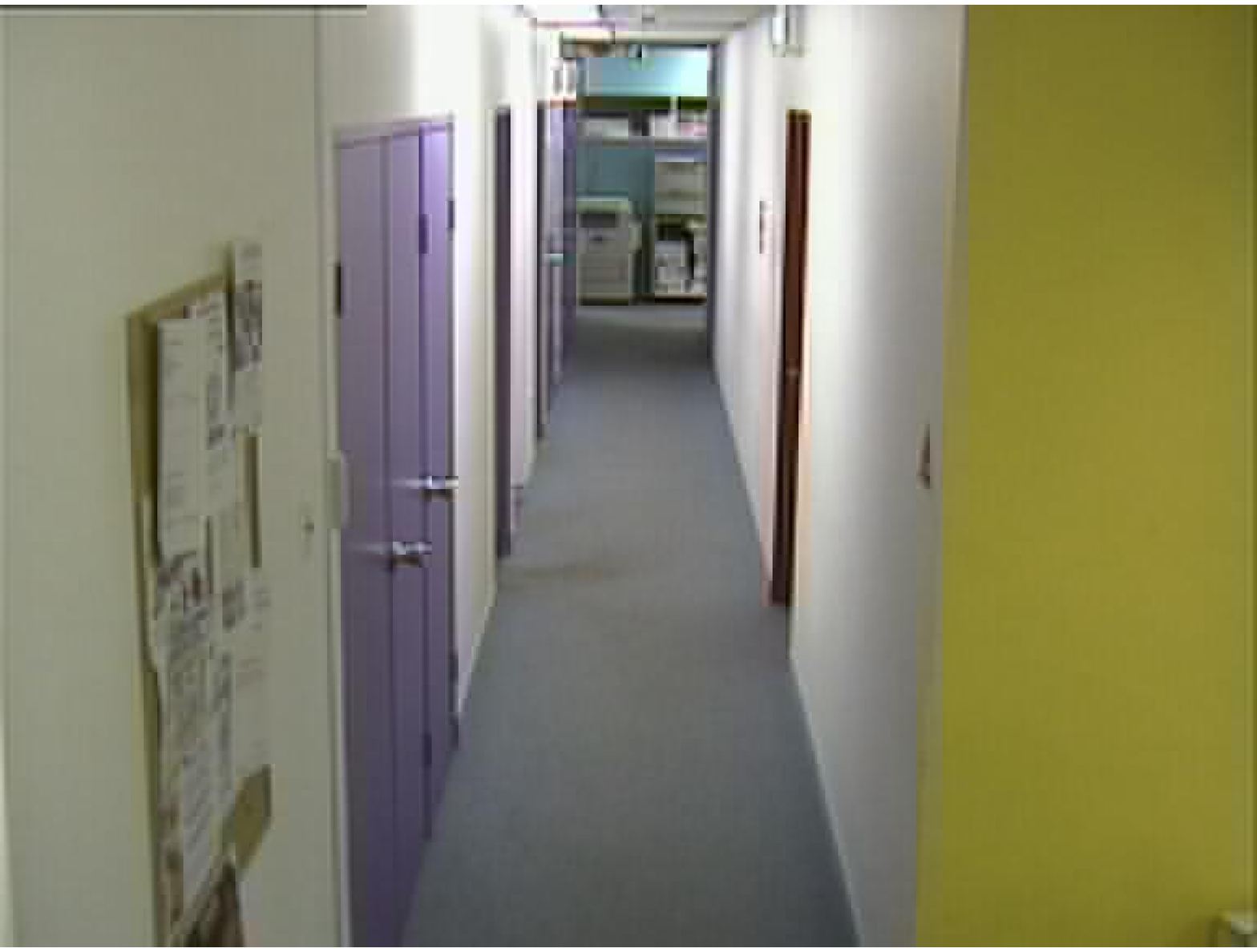} & \includegraphics[width=0.3\textwidth]{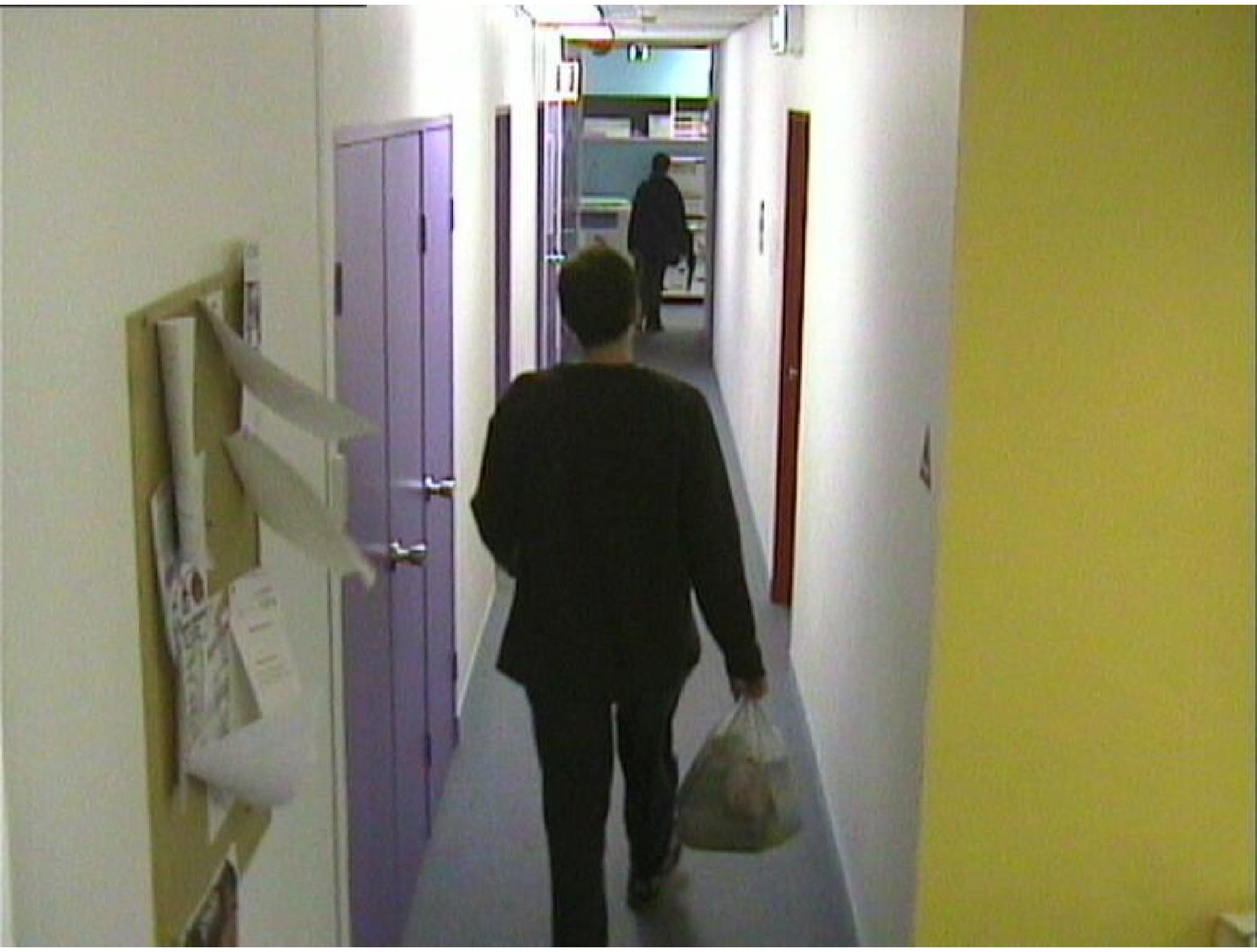} & \includegraphics[width=0.3\textwidth]{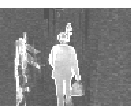}\tabularnewline
\end{tabular}

\caption{Values for $\kappa$ for an example image. Scene model (left), input
image(center), classifier output (dark = matcibh, bright = no match)
(right)\label{fig:DCTclassifierResult-values-for}}

\end{figure*}

As an initial mode matching value, a weighted sum of the absolute
differences between each of the coefficients of block $B$ and mode
model $M_{m}$ is calculated as follows:\begin{equation}
\kappa(B,M_{m})=\overset{_{7}}{\underset{_{i=0}}{\sum}}a_{i}\left|c_{i}\right|\label{eq:SumOfCoeffs}\end{equation}
where $c_{i}$ is the difference between the i-th coefficient of the
input block and the i-th coefficient of the mode model $M_{m}$and
$a_{i}$is the trained weight. An example of $\kappa(B,M_{m})$ values
for an image is shown in Figure \ref{fig:DCTclassifierResult-values-for}.

Trained weight $a_{i}$ is determined as follows. Naive Bayes training
is used to determine the probability that a given set of input coefficients
match a given set of mode coefficients based upon a set of training
data. A logistic regression is then applied to determine coefficients
$a_{0}$...$a_{7}$. This yields $a_{i}=$the linear regression coefficient
of $log(P(match|c_{i})/P(nomatch|c_{i}))$ for all $i$.

\begin{figure}
\includegraphics[width=1\columnwidth]{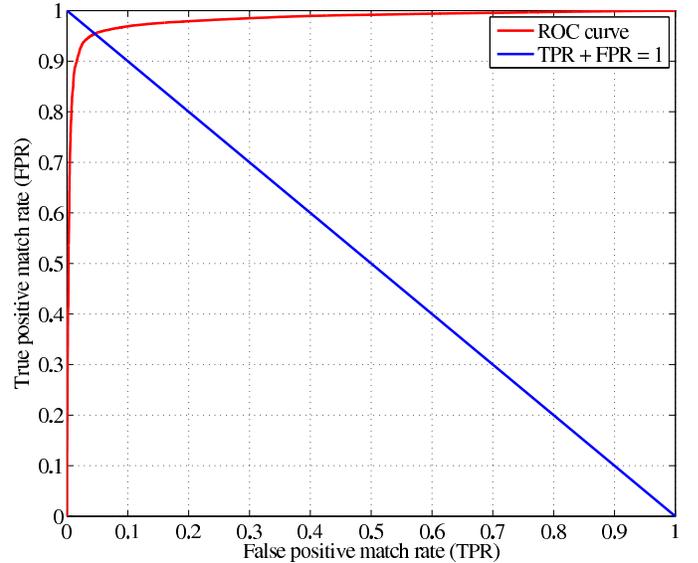}

\caption{ROC curve of DCT blocks for match versus no match classification\label{fig:ROC-curve-of}}

\end{figure}

The threshold $T$ for creating a new mode must also be determined.
That is, the threshold that determines when a probability is not strong
enough to warrant a mode match. We use the criterion \{true positive
rate + false positive rate = 1\} to determine the threshold level
for $\kappa(B,M_{m})$ where P(mode match) = 0.5. This can be determined
and visualized by using a Receiver Operating Characteristic (ROC)
graph (Figure \ref{fig:ROC-curve-of}). This method is also used for
spatial training as detailed in section \ref{sub:Spatial-classifier}.

\subsection{\label{sub:Spatial-classifier}Spatial classifier}

Video sequences from network cameras often contain noise from camera
refocusing, lighting changes and sensor noise. To avoid accumulation
of errors in the scene model, a spatial classifier is included in
this system. The spatial classifier takes not only the current block
into account, but also its neighbors. The underlying assumption is
that for a given DCT block at a given point in time in an image sequence,
a mode is more likely to be a match if the adjacent DCT blocks match
modes that were created at a similar time to when that mode was created. 

\begin{figure}
\begin{tabular}{cc}
\includegraphics[width=0.45\columnwidth]{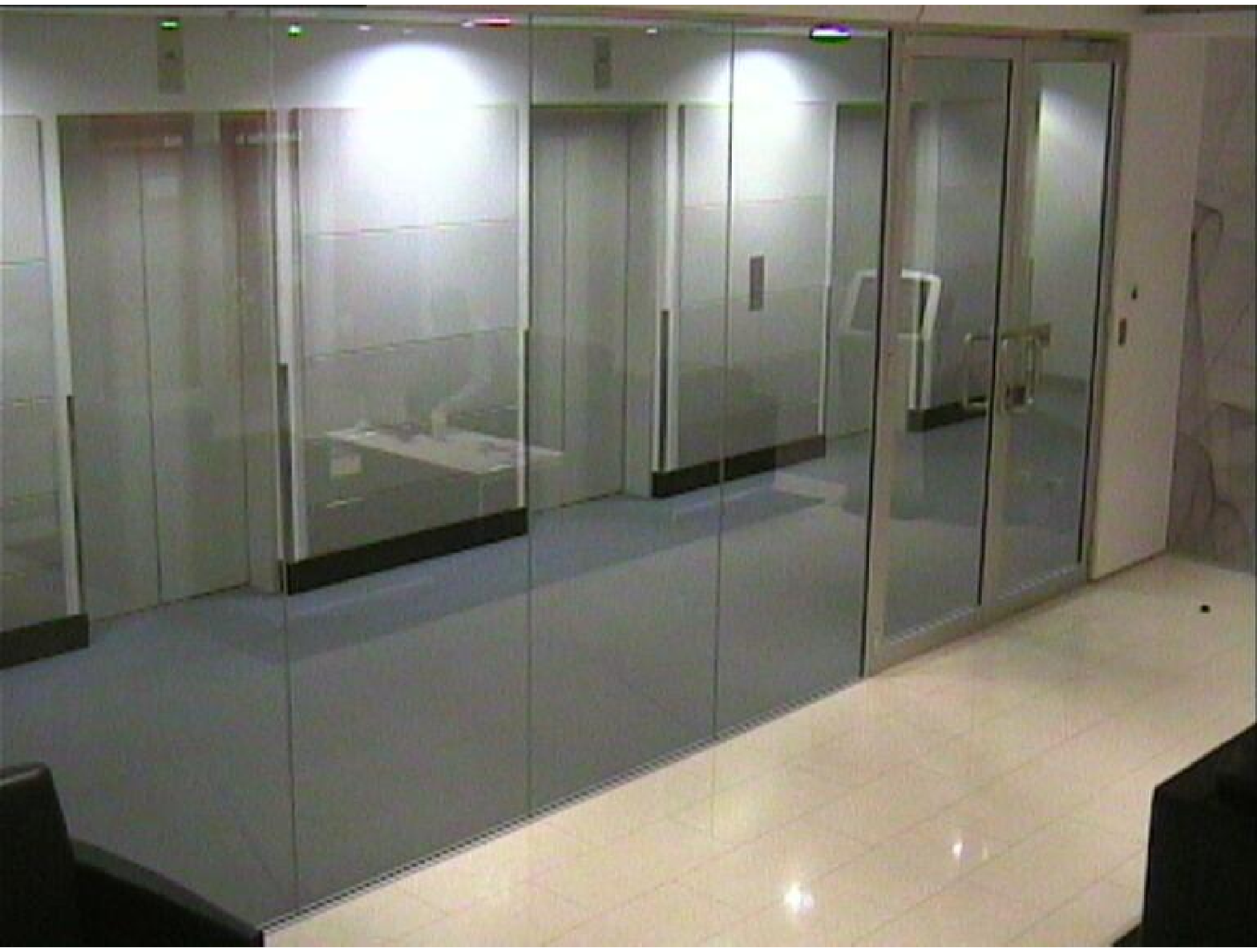} & \includegraphics[width=0.45\columnwidth]{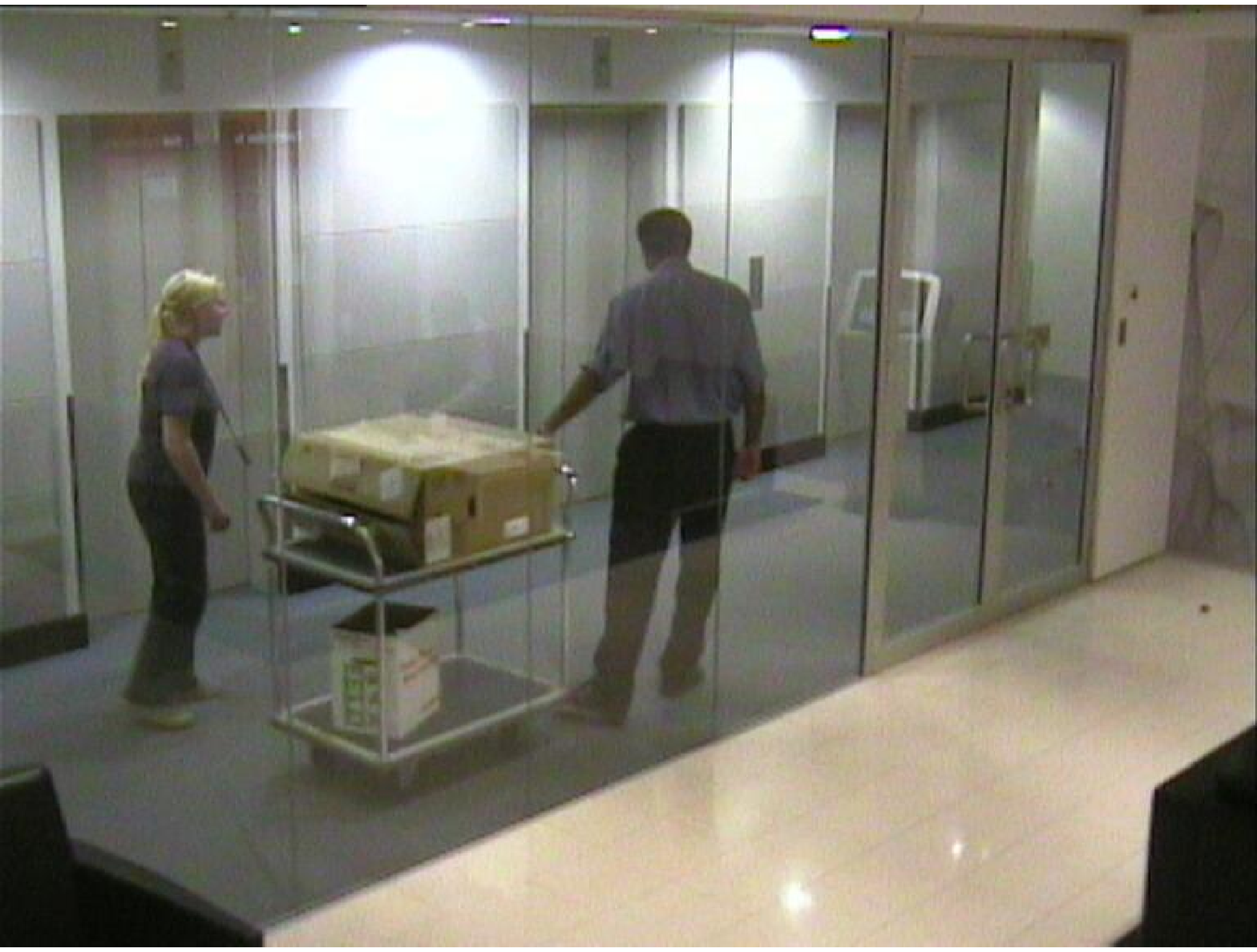}\tabularnewline
\includegraphics[width=0.45\columnwidth]{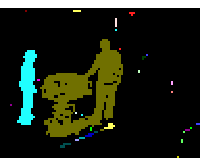} & \includegraphics[width=0.45\columnwidth]{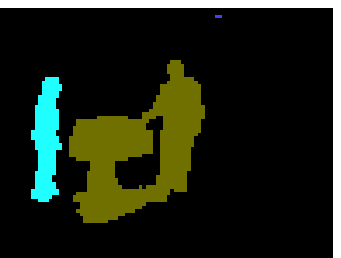}\tabularnewline
\end{tabular}

\caption{Indoor scene: Empty scene (top left), system input (top right), system
output for 0 spatial iterations (bottom left), system output for 3
spatial iterations (bottom right)\label{fig:Indoor-scene:-Empty}}

\end{figure}

Accumulation of noise in the scene model is particularly detrimental
to age information. We model the creation frame and hit count of each
mode in each block of the sequence. Incorrect mode matching causes
the age information to be invalid. These errors can propagate with
time due to mode model updating and mode removal, see figure \ref{fig:Indoor-scene:-Empty}
(bottom left) for an example. The spatial classifier corrects the
DCT classifier result for each block in the sequence by using information
from adjacent blocks.

Let adjacent DCT blocks to a given DCT block in an image be defined
using 4-connectivity as the blocks directly above, below, left and
right of the given DCT block. For each mode model $M$ in the scene
model for a given DCT block, $A$ is the number of mode models of
adjacent blocks that are temporally similar to $M$. Temporal similarity
is defined according to the difference in creation time between $M$and
the adjacent block's mode model, using threshold t. Value $A$ affects
the likelihood that $M$ is the best match. 

\begin{figure*}
\begin{tabular}{ccc}
\includegraphics[width=0.3\textwidth]{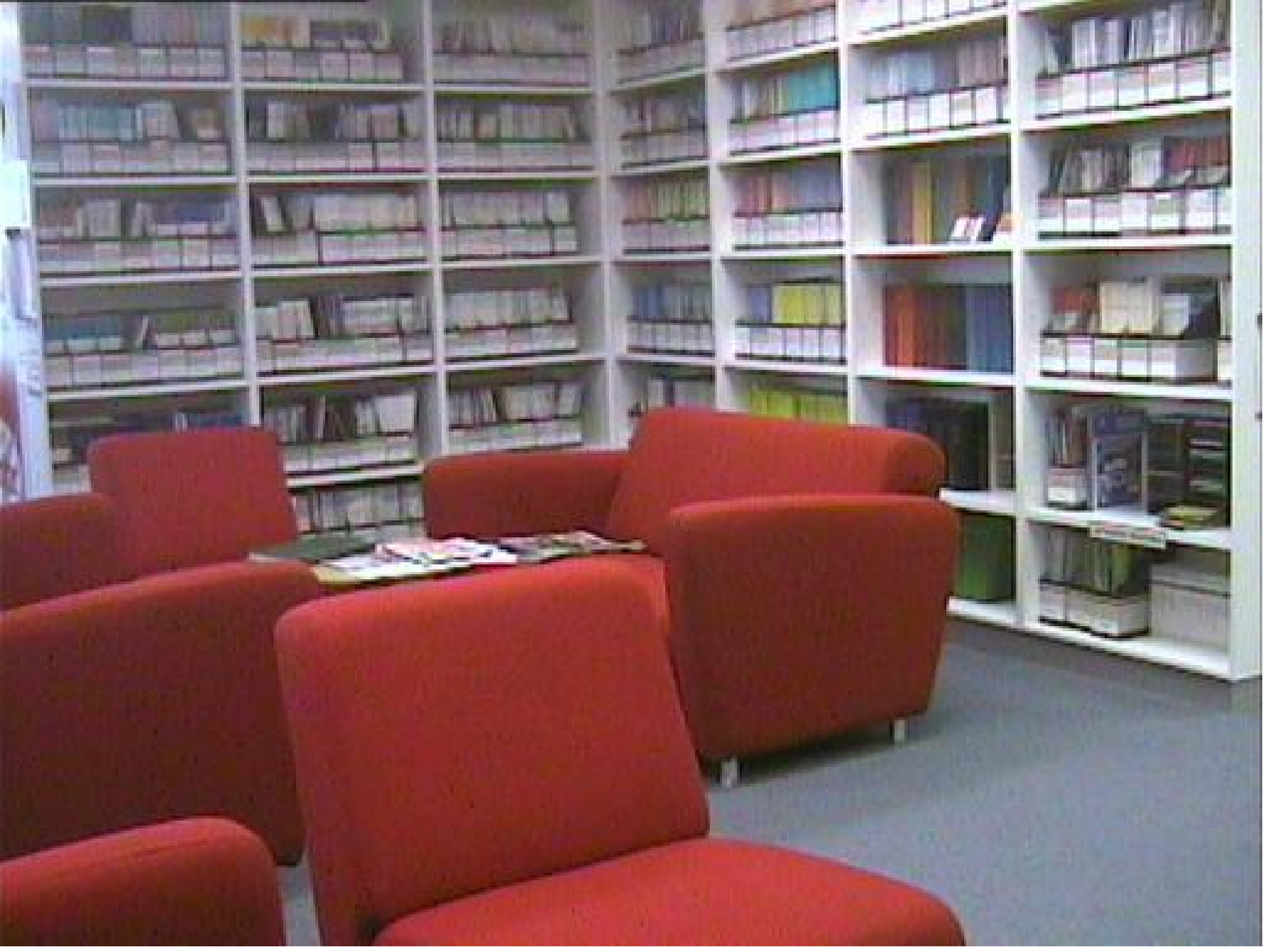} & \includegraphics[width=0.3\textwidth]{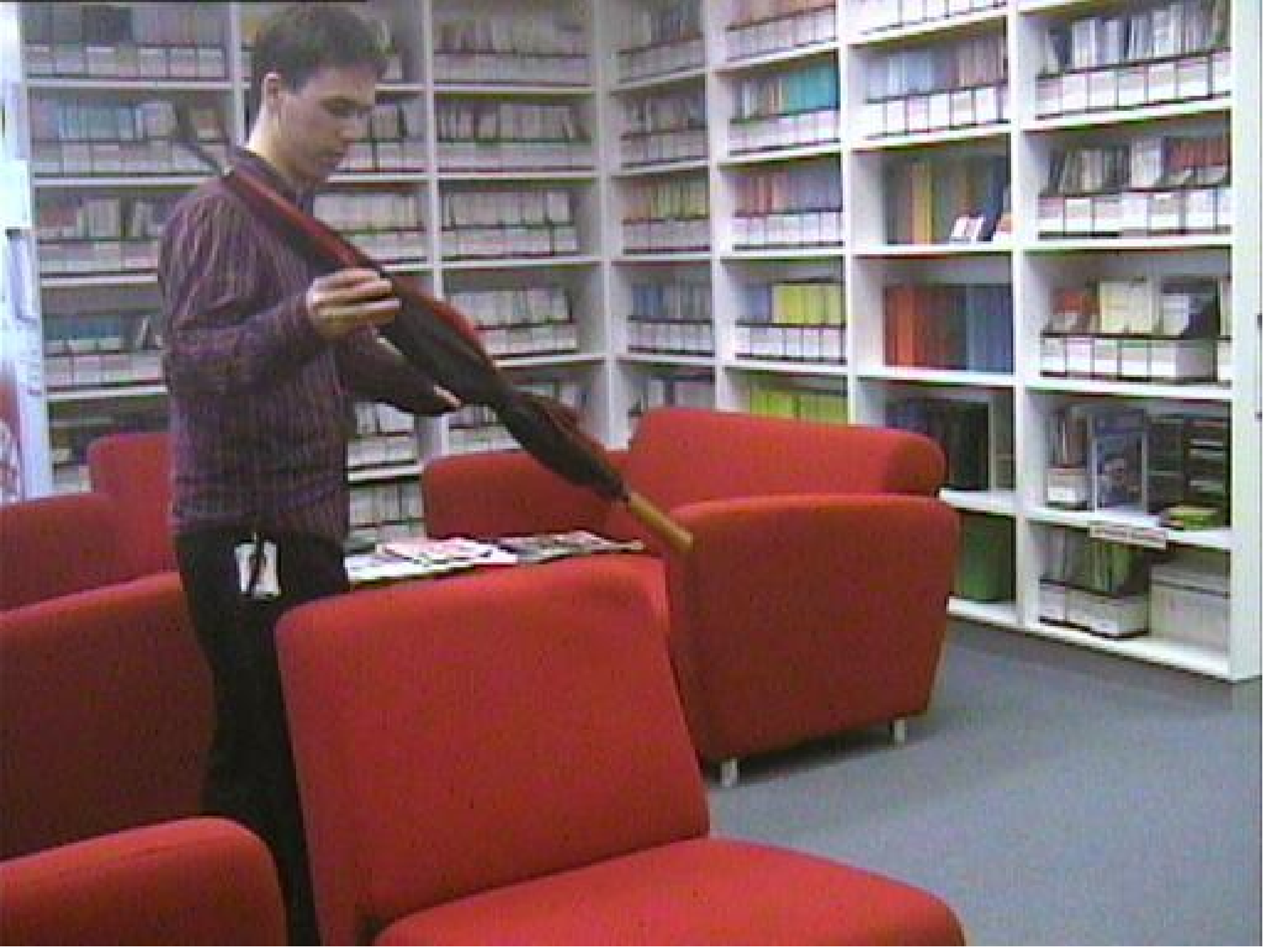} & \includegraphics[width=0.3\textwidth]{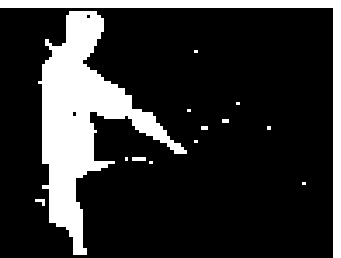}\tabularnewline
\includegraphics[width=0.3\textwidth]{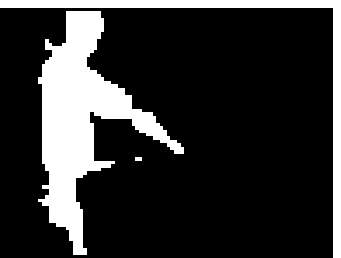} & \includegraphics[width=0.3\textwidth]{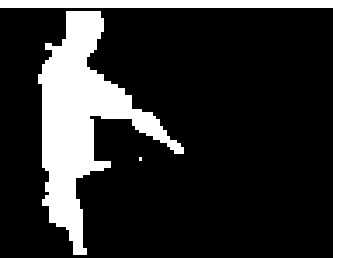} & \includegraphics[width=0.3\textwidth]{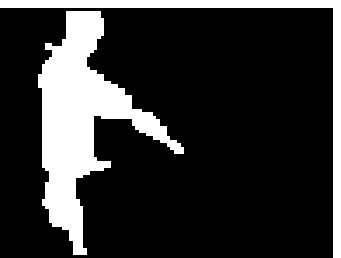}\tabularnewline
\end{tabular}

\caption{Training sequence images, from left to right: Scene model (top left),
foreground (top), 0 spatial iterations (top right), 1 spatial iteration
(bottom left), 2 spatial iterations (bottom), 3 spatial iterations
(bottom right) \label{fig:Training-sequence}}

\end{figure*}

The final mode match value for a block is $\kappa(B,M_{m})$ + $\lambda(I,A)$,
where $\lambda(I,A)$ accesses a lookup table for the $I$-th iteration
and the number of similar neighbors $A$. The lookup table is constructed
during an offline training stage. For each possible value of $M$
and $I$, an ROC graph is determined and $\lambda(I,A)$ is taken
at the point on the graph where the sum of the true positive rate
and false positive rate equals 1. This method is similar to the method
for determining the threshold level for the DCT classifier described
in section \ref{sub:DCT-classifier}. An example of one of the training
sequences, at different stages of iteration is shown in figure \ref{fig:Training-sequence}.

\subsection{\label{sub:Active-mode-bonus}Active mode bonus}

Mode persistence is used to improve classification. While some object
detection applications focus on tracking moving objects, other applications
have a greater need for stable and consistent detection of stationary
objects. By including a probability measure \textcolor{black}{that}
is added to the match probability for modes seen within the last few
frames, this trade-off can be adjusted by users of the system. Low
(or zero) contributions from mode persistence result in better detection
of moving objects. Increasing the mode persistence probability results
in more stable and consistent stationary object detection, which also
reduces the impact of noise eroding a stationary object.

\subsection{\label{sub:Approximated-median-filter}Approximated median filter
scene model updating}

\begin{figure}
\includegraphics[width=1\columnwidth]{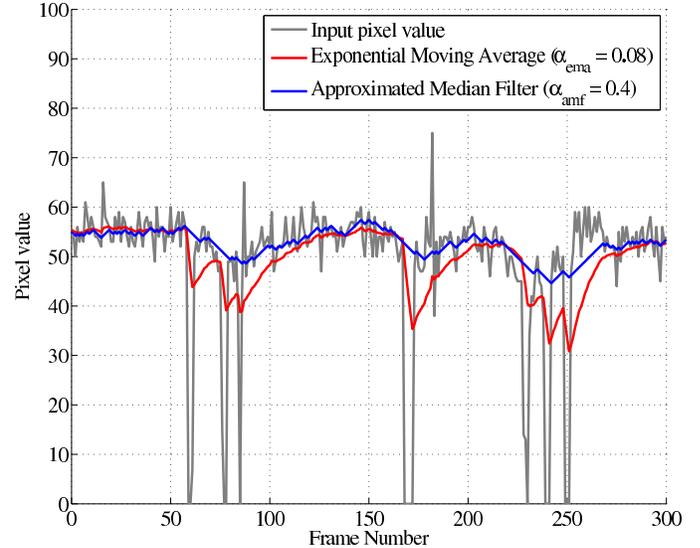}

\caption{Exponential Moving Average versus Approximated Median Filter\label{fig:Exponential-Moving-Average}}

\end{figure}

The commonly used method of using an exponential moving average (EMA)
to model scene modes \cite{Stauffer00,Li01} suffers from a number
of problems. There is a noise increase in very dark and very light
regions, due to sensor noise, lighting changes and foreground objects
perturbing the model. Further, the amount of time required to return
to the original state after an impulse signal is dependent upon the
size of the impulse. In addition, the method is computationally expensive
due to the multiplications and precision required.

We use an approximated median filter (AMF) \cite{McFarlane95} to
address the aforementioned issues:

If $x_{n}>y_{n}+\alpha_{amf}:y_{n+1}=y_{n}+\alpha_{amf}$

If $x_{n}>y_{n}-\alpha_{amf}:y_{n+1}=y_{n}-\alpha_{amf}$

If $x_{n}\leq y_{n}+\alpha_{amf}$ and $x_{n}\geq y_{n}-\alpha_{amf}:y_{n+1}=x_{n}$ 

where $x_{n}$is the input coefficient value, $y_{n}$is the mode
model coefficient value, and $\alpha_{amf}$ is the learning parameter
that determines how fast the mode model adapts. 

\begin{figure}
\includegraphics[width=1\columnwidth]{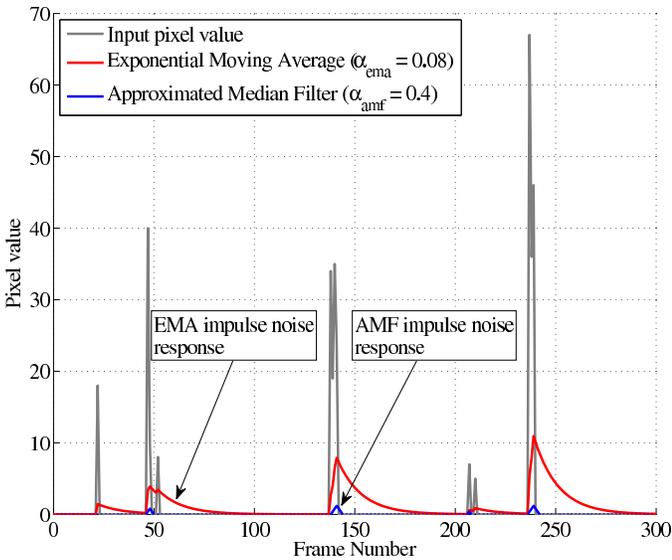}

\caption{Response to impulse noise for a dark background pixel \label{fig:EMAdark}}

\end{figure}

Biased noise is decreased in bright and dark regions, since the recovery
time is dependent upon the duration of the noise signal but not upon
the intensity of the noise, see figure \ref{fig:EMAdark} for an example.
Recovery time for impulsive noise is 1 frame, unlike the exponential
moving average.

\subsection{Mode removal}

Mode models have to be removed for two reasons. The first reason is
practical: a system may run out of memory if too many modes have been
created. This is especially important in systems where the system
is allocated a fixed maximum amount of memory, e.g. in the context
a bigger system where a large number of cameras is supported. In addition,
more modes means more processing power is needed. A maximum number
of modes may be introduced to make the system performance feasible
and predictable. The second reason is regardless of the availability
of system resources. Modes must be removed from the system in order
to reduce the probability of new objects being matched to unrelated
mode models. 

\begin{figure*}
\begin{tabular}{ccc}
\includegraphics[width=0.3\textwidth]{caseL695_bgm} & \includegraphics[width=0.3\textwidth]{caseL695_FG} & \includegraphics[width=0.3\textwidth]{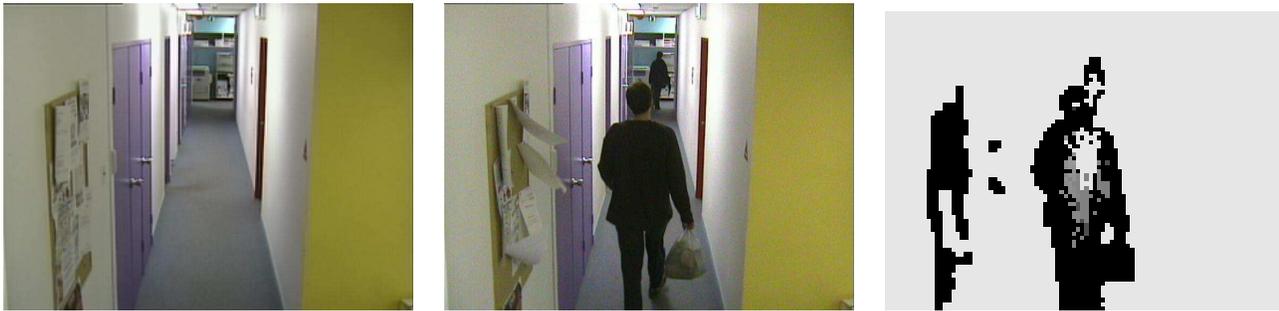}\tabularnewline
\end{tabular}

\caption{Modes that should have been deleted earlier from frequent traffic
in the hallway. Scene model (left), input image(center), age image
(black = new, green = old), (right)\label{fig:AgeDeletion}}

\end{figure*}

Determining when to remove a mode is a trade-off decision. If modes
are removed too soon, objects that are occluded will not match the
scene model when they reappear. This is known as the revealed background
problem. Modes that are not removed quickly enough have an increased
likelihood of being matched to unrelated objects, see for example
figure \ref{fig:AgeDeletion}. We remove a mode model when it hasn't
been matched to an incoming block for a number of frames. For each
mode model, a potential removal frame is computed when the mode model
is created. The value is updated each time the mode model is matched
to an incoming block. The mode removal frame $f_{mr}$ is computed
as follows:

$f_{mr}=f_{current}+C_{s}+C_{v}*HitCount$,

where $C_{s}$is a constant indicating the minimum survival time of
the mode model, and $C_{v}$is a constant indicating the minimum percentage
of time this part of the scene must remain visible to retain its temporal
information.

Each time a new frame is processed, the $f_{mr}$ value of a mode
model is checked. If the value matches the current frame number $f_{current}$,
the mode model is removed regardless of the total number of mode models
for the block. Note that always at least one mode model remains if
$C_{s}$ is greater than 0. In addition, each time a new mode model
is created, and the maximum number of modes per DCT block has been
reached, the mode with the earliest mode removal frame value is removed.

\section{\label{sec:Evaluation}Evaluation}

\subsection{Experimental setup}

An evaluation video data set was recorded using a network camera providing
a Motion-JPEG stream (resolution 768x576). Although the SAMMI algorithm
is not specific to any particular flavor of JPEG compression, by using
a specific camera we have full control over the recording circumstances,
and we avoid any influence of video transcoding. The evaluation data
set targets difficult, yet realistic scenarios. That is, the data
set is typically not representative for the content an object detection
system may encounter in practice. The focus of our evaluation is on
indoor scenarios, where there is artificial lighting with in some
cases additional natural lighting through windows. The data set is
biased towards stationary foreground objects, stressing the importance
of abandoned object detection and object removal applications. Each
video starts with an empty scene to allow for initialization of the
background model. See table \ref{tab:description_videos} for details
on the videos. The screen savers present in some of the videos serve
to stretch the modeling in that area of the scene for research purposes,
as they have more states than the maximum number of modes.

For each of the 10 videos a bounding box ground truth was created.
In addition, for more precise evaluation for in total 10,185 frames
in 6 videos, an object mask ground truth was created, which is expected
to be accurate within an error margin of 4 pixels. Note that the creation
of ground truth is a problem in itself, and that different people
or the same person at a different time may segment the same video
differently. As creation of the precise object mask ground truth is
very expensive manual labor, it was done for a subset of the videos
only.

The DCT classifier and spatial classifier are trained on 12 frames
acquired from 7 videos. The scenes and locations at which the training
videos were recorded are different from the scenes used in the test
set.

\begin{table*}
\begin{tabular}{|>{\raggedright}p{0.44\textwidth}|>{\raggedright}p{0.41\textwidth}|>{\raggedright}p{0.06\textwidth}|}
\hline 
Description & Difficulty & Length (frames) \tabularnewline
\hline
\hline 
1. Several people walk through reception area.  & Screen saver. Low contrast difference between some objects and background.  & 3439 {[}896]\tabularnewline
\hline 
2. Person sits down to read newspaper. & Stationary objects. Screen saver. Low contrast difference between
some objects and background.  & 1711\tabularnewline
\hline 
3. Person sits down to read, shuffles magazines, leaves a small object
behind, and returns to fetch it. & Stationary objects. Screen saver. Low contrast difference between
some objects and background.  & 6600 {[}6299]\tabularnewline
\hline 
4. Person moves around trolley in reception. There is a visible computer
screen. & Structural changes of background. & 2939 {[}700]\tabularnewline
\hline 
5. Person sits in office, moves existing chairs around.  & Illumination change. Stationary objects. Screen saver.  & 1658\tabularnewline
\hline 
6. Person sits in office, moves existing chairs around. Window blinds
are swaying. & Rapid illumination change. Stationary objects.  & 4159\tabularnewline
\hline 
7. Person sits down, deposits object, moves existing papers. & Structural changes of background. Screen saver.  & 1435 {[}572]\tabularnewline
\hline 
8. Adult and child walk in corridor, from far away towards the camera. & Low contrast difference between some objects and background. & 448 {[}370]\tabularnewline
\hline 
9. Multiple people walk through hallway, leave objects behind, open/close
boxes on the scene. & Stationary objects. & 2117 \tabularnewline
\hline 
10. Several people walk through corridor, enter/exit at various points,
stand still, abandon small objects. & Stationary objects. & 1984 {[}1348]\tabularnewline
\hline
\end{tabular}

\caption{Description of videos used in the test set. The number of frames used
in object mask ground truth is given in square brackets.\label{tab:description_videos}}

\end{table*}

\subsection{Evaluation criteria}

Like other object detection algorithms, the SAMMI algorithm has general
applicability. Whether the produced output is good in a relative or
absolute sense depends on the context in which it is used. The requirements
for object detection in an intruder alert system are very different
from those in a people counting application. Similarly, a system that
alerts a security guard will give a higher penalty to false alarms
than a system that does event-based recording. We evaluate the system
output at two levels:

\begin{itemize}
\item Pixel-level accuracy: exact matching between ground truth object masks
and the detected objects. Note that the output of the SAMMI method
has a coarser granularity than pixel, viz. 8x8 blocks. Hence, it is
not possible for our method to score the maximum on this level, while
pixel-based algorithms could theoretically reach a score of 100\%.
Also, the problem of inconsistency in ground truths mentioned before
may not even allow a perfect segmentation algorithm to score 100\%.
\item Tracking suitability: the impact of the segmentation results on a
tracker application.
\end{itemize}
For pixel-level accuracy, we determine a true positive count by an
AND operation on the detection result and the ground truth. Similarly,
false positives and false negatives are found. An F1-measure (harmonic
mean of recall and precision) is then computed. 

For tracking suitability, the number of detected blobs is set out
against the number of blobs that could be related to a bounding box.
Only one blob is associated per ground truth bounding box, resulting
in a penalty for fragmentation of objects. Because of the simplicity
of the measure, it does not allow us to draw any conclusions about
the absolute results. However, the measure does tell to what extent
fragmentation has been reduced. Although tracking applications may
be able to deal with fragmentation, lower fragmentation leads to lower
computational demands and decreases the chance of tracking spurious
objects. The tracking suitability measure should always be used in
combination with another object detection measure, such as the pixel-level
F1-measure, as it does not penalize for missed detections.

Finally, we measure the computational expense and the memory usage
of the scene model. As the foremost purpose of the SAMMI method is
an increase in speed, it is important to measure the impact on resource
usage. Although actual implementations of the evaluated methods have
not been optimized for speed, processing time still gives an indication
of the difference between methods. Experiments were run on one 2.16
GHz Intel Core Duo processor, and measurements include file I/O. Memory
usage is estimated for the persistent scene model only, that is the
memory usage in between processing 2 frames.

\subsection{Results}

\begin{figure}
\includegraphics[width=1\columnwidth]{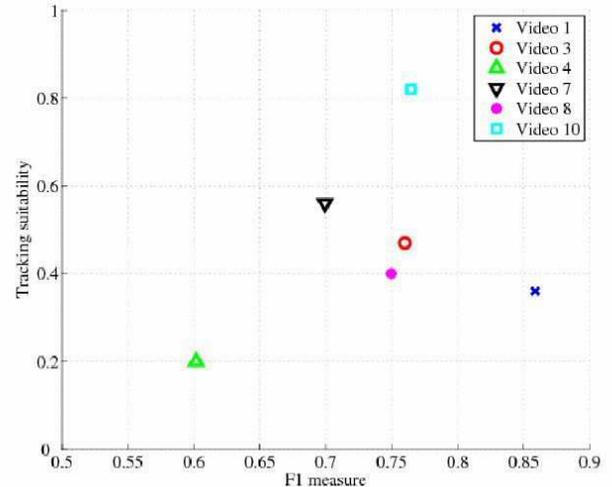}

\caption{SAMMI performance, using 3 iterations, at pixel level and tracking
level for videos for which an object mask ground truth is available.\label{fig:per-video-results}}

\end{figure}

We compare the SAMMI object detection method for different numbers
of iterations to the Mixture of Gaussians method. For SAMMI, the maximum
number of mode models was set to 5. Results are shown in table \ref{tab:performance}.
Per video, results for SAMMI using 3 iterations are shown in figure
\ref{fig:per-video-results}

\begin{table*}
\begin{tabular}{|l||>{\centering}p{1.7cm}|>{\centering}p{1.7cm}|>{\centering}p{1.7cm}|>{\centering}p{1.7cm}|>{\centering}p{1.7cm}|}
\hline 
\hfill{}Method & Mixture of Gaussians & \multicolumn{4}{c|}{SAMMI}\tabularnewline
\cline{3-6} 
\hline 
\hfill{}Number of iterations & - & 0 & 1 & 2 & 3\tabularnewline
\hline
\hline 
F1 & 0.55 & 0.79 & 0.80 & 0.80 & 0.80\tabularnewline
\hline 
Tracking suitability & 0.38 & 0.27 & 0.31 & 0.34 & 0.38\tabularnewline
\hline 
Processing time (frames per second) & 5 & 30 & 29 & 27 & 25\tabularnewline
\hline 
Scene model memory usage (KB) & 36,288 & 1,080 & 1,080 & 1,080 & 1,080\tabularnewline
\hline
\end{tabular}

\caption{Results for methods according to the evaluation measures.\label{tab:performance}}

\end{table*}

The Mixture of Gaussian method suffers from the trade-off between
detecting moving objects and detecting stationary objects. Although
it has a precision higher than SAMMI (0.79 versus 0.69), its recall
is significantly lower (0.42 versus 0.95) indicating a large number
of false negatives. The difference in performance at pixel level is
not significant for the various iterations. However, the larger number
of iterations shows a significant increase in tracking suitability,
corresponding to a decrease in object fragmentation.

Since our experimental setup does not focus on measuring processing
time precisely, it is inappropriate to draw final conclusions about
the difference between the methods. However, the significant difference
between Mixture of Gaussians and our method supports the expectation
that our method is significantly faster. Similarly, the difference
in memory usage for scene modeling is significant. 

\section{\label{sec:Conclusions}Conclusions}

Computationally inexpensive background modeling can be done without
a significant penalty in accuracy. The use of DCT information without
transforming image information to the pixel domain still allows for
good accuracy while making significant savings in resource usage.
The use of a fast approximated median method makes the modeling robust
to noise in bright and dark regions of a scene, while it is faster
than the conventional exponential moving average approach. Fragmentation
noise is reduced by several iterations of neighbor adapted classification
based on temporal coherency of objects.

Another advantage of the SAMMI system is its configurability. Users
can configure the trade-off between detecting new moving objects and
existing stationary objects using the active mode bonus. Similarly,
users can make trade-offs for removing modes by specifying the minimum
percentage of time a part of the scene must remain visible to retain
its temporal information. 

The spatial processing outlined in this paper allows for a greater
variability in the size of objects, particularly small objects, that
can be successfully detected. The filtering of local noise in the
image sequence that would otherwise cause spurious blobs to be detected
is embedded within the scene modeling process.

Through low resource usage while preserving acceptable accuracy, the
lightweight object detection method presented in this paper increases
the feasibility of deploying video analysis systems in the real world. 

\bibliographystyle{ieee}
\bibliography{mm_arxiv}

\newpage{}
\end{document}